# Deep-Net: Deep Neural Network for Cyber Security Use Cases

Vinayakumar R[1], Barathi Ganesh HB[1,2], Prabaharan Poornachandran[3], Anand Kumar M[4] and Soman KP[1]

[1]Centre for Computational Engineering and Networking (CEN), Amrita School of Engineering, Coimbatore

[2]Arnekt Solutions Pvt Ltd, Pune, Maharashtra, India

[3]Center for Cyber Security Systems and Networks, Amrita School of Engineering, Amritapuri, Amrita Vishwa Vidyapeetham, India

[4]Department of Information Technology, National Institute of Technology, Karnataka, Surathkal. Mangalore.

Email: vinayakumarr77@gmail.com, barathiganesh.hb@arnekt.com

## Abstract

Deep neural networks (DNNs) have witnessed as a powerful approach in this year by solving long-standing Artificial intelligence (AI) supervised and unsupervised tasks exists in natural language processing, speech processing, computer vision and others. In this paper, we attempt to apply DNNs on three different cyber security use cases: Android malware classification, incident detection and fraud detection. The data set of each use case contains real known benign and malicious activities samples. The efficient network architecture for DNN is chosen by conducting various trails of experiments for network parameters and network structures. The experiments of such chosen efficient configurations of DNNs are run up to 1000 epochs with learning rate set in the range [0.01-0.5]. Experiments of DNN performed well in comparison to the classical machine learning algorithms in all cases of experiments of cyber security use cases. This is due to the fact that DNNs implicitly extract and build better features, identifies the characteristics of the data that lead to better accuracy. The best accuracy obtained by DNN and XGBoost on Android malware classification 0.940 and 0.741, incident detection 1.00 and 0.997 fraud detection 0.972 and 0.916 respectively.

# 1 Introduction

In this era of technical modernization, explosion of new opportunities and efficient potential resources for organizations have emerged but at the same time these technologies have resulted in threats to the economy. In such a scenario proper security measures plays a major role. Now days', hacking has become a common practice in organizations in order to steal data and information. This highlights the need for an efficient system to detect and prevent the fraudulent activities. Cyber security is all about the protection of systems, networks and data in the cyberspace.

Malware remains one of the maximum enormous security threats on the Internet. Malware are the software's which indicate malicious activity of the file or programs. These are unwanted programs since they cause harm to the intended use of the system by making it behave in a very different manner than it is supposed to behave. Solutions with Antivirus and blacklists are used as the primary weapons of resistance against these malwares. Both approaches are not effective. This can only be used as an initial shelter in real time malware detection system. This is primarily due to the fact that both approaches are completely fails at detecting the new malware that is created using polymorphic, metamorphic, domain flux and IP flux.

Machine learning algorithms have played a pivotal role in several use cases of Cyber security [1]. Fortunately, deep learning approaches are prevailing subject in recent days due to the remarkable performance in various long-standing artificial intelligence (AI) supervised and unsupervised challenges [2]. The efficacy of deep learning architectures are transformed to various use cases of cyber security [25], [26], [27], [28], [29], [30], [31], [32], [33], [34], [35], [36], [37], [38], [39], [40]. This paper evaluates the effectiveness of deep neural network (DNN) for cyber security use cases: Android malware classification, incident detection and fraud detection.

The paper is structured as follows. Section 2 discusses the related work. Section 3 discusses the background knowledge of deep neural network (DNN). Section 4 presents the proposed methodology including the description the data set. Results are displayed in Section 5. Conclusion is placed in Section 6.

## 2 Related work

This section discusses the related work for cyber security use cases: Android malware classification, incident detection and fraud detection.

Static and dynamic analysis is the most commonly used approaches in Android malware detection [3]. In static analysis, android permissions are collected by unpacking or disassembling the app. In dynamic analysis, the run-time execution characteristics such as system calls, network connections, power consumption, user interactions and memory utilization. Mostly, commercial systems use combination of both the static and dynamic analysis. In Android devices, static analysis is preferred due to the following advantageous such as less computational cost, low resource utilization, light-weight and less time consuming. However, dynamic analysis has the capability to detect the metamorphic and polymorphic malwares. In [4] evaluated the performance of traditional machine learning classifiers for android malware detection with using the permission, API calls and combination of both the API calls and permission as features. These 3 different feature sets were collected from the 2510 APK files. All traditional machine learning classifiers performance is good with combination of API calls and permission feature set in comparison to the API calls as well as permission. [5] proposed MalDozer that use sequences of API calls with deep learning to detect Android malware and classify them to their corresponding family. The system has performed well in both private and public data sets, Malgenome, Drebin.

Recently, the privacy and security for cloud computing is briefly discussed by [6]. The discussed various 28 cloud security issues and categorized those issues into five major categories. [7] proposed machine learning based anomaly detection that acts on different layers e.g., the network, the service, or the workflow layers. [8] discussed the issues in creating the intrusion detection for the cloud infrastructure. Also, how rule based and machine learning based system can be combined as hybrid system is shown. [9] discussed the security problems in cloud and proposed incident detection system. They showed how incident detection system can perform well in comparison to the intrusion detection.

In [10] did comparative study of six different traditional machine learning classifiers in identifying the financial fraud. In [11] discussed the applicability of data mining approaches for financial fraud detection.

## 3 Background

The purpose of this section is to discuss the concepts of deep neural network (DNN) architecture concisely and promising techniques behind to train DNN.

Artificial neural networks (ANNs) represent a directed graph in which a set of artificial neuron generally called as units in mathematical model that are connected together with edges. This influenced by the characteristics of biological neural networks, where nodes represent biological neurons and edges represent synapses. A feed forward network is a type of ANNs.

A feed forward network (FFN) consists of a set of units that are connected together with edges in a single direction without formation of a cycle. They are simple and most commonly used algorithm. Multi-layer perceptron (MLP) is a subset of FFN that consist of 3 or more layers with a number of artificial neurons, termed as units. The 3 layers are input layer, a hidden layer and output layer. There is a possibility to increase the number of hidden layers when the data is complex in nature. So, the number of hidden layer is parameterized and relies on the complexity of the data. These units together form an acyclic graph that passes information or signals in forward direction from layer to layer without the dependence of past input. MLP can be written as $O: \mathbb{R}^p \times \mathbb{R}^q$ where $p$ and $q$ are the size of the input vector $x = x_1, x_2, \cdots, x_{p-1}, x_p$ and output vector $O(x)$ respectively. The computation of each hidden layer $Hl_i$ can be mathematically formulated as follows.

$$HI_i(x) = f(w_i^T x + b_i)$$

$$HI_i : \mathbb{R}^{d_{i-1}} \to \mathbb{R}^{d_i}$$

$$f : \mathbb{R} \to \mathbb{R}$$

$w_i \in \mathbb{R}^{d \times d_{i-1}}$, and $b \in \mathbb{R}^{d_i}$ $f$ is an element wise non-linearity function. This can be either *logistic sigmoid* or *hyperbolic tangent function*. *logistic sigmoid* has value either 0 or 1 whereas [1,-1] range of values for *hyperbolic tangent function*. If we want to use

MLP for multi class classification problem, then the output usually have multiple neurons. For this, *soft* max function can be used. This provides the probabilities of each class and selecting the highest one results in crisp value.

$$sigmoid = \sigma(x) = \frac{1}{1+e^{-z}}$$

$$hyperbolic \tan gent = \tanh(z) = \frac{e^{2z}-1}{e^{2z}+1}$$

$$SF(Z)_i = \frac{e^{z_i}}{\sum_{j=1}^{n} e^{z_j}}$$

When then network consist of $l$ hidden layers then the combined representation of them can be generally defined as

$$HI(x) = HI_l(HI_{l-1}(HI_{l-2}(\cdots(HI_1(x)))))$$

This way of stacking hidden layers on top of each other is typically called as deep neural network (DNN) Each hidden layer uses ReLU as non-linear activation function. This helps to reduce the state of vanishing and error gradient issue [12, 13, 14].

## 3.1 Rectified Linear Unit (ReLU)

Rectified linear units (ReLU) have been turned out to be more proficient and are capable of accelerating the entire training process altogether [12]. Selecting $ReLU$ is a more efficient way when considering the time cost of training the vast amount of data. The reason being that not only does it substantially speeds up the training process but also possesses some advantages when comparing to the traditional activation function including logistic function and hyperbolic tangent function [13]. We refer to neurons with this nonlinearity following [14].

## 4 Experiments

We consider TensorFlow [15] in conjunction with Keras [16] as software framework. To increase the speed of gradient descent computations of deep learning architectures, we use with GPU enabled TensorFlow in single NVidia

GK110BGL Tesla k40. All deep learning architectures are trained using the back propagation through time (BPTT) technique.

## 4.1 Description of Data sets

Task 1 (Android Malware Classification) data set includes 37,107 unique API information from 61,730 APK files [17]. These APK (application package) files were collected from the Opera Mobile Store over the period of January to September of 2014. When a user runs an application, a set of APIs will be called. Each API is related to a particular permission. The execution of the API may solely achieve success within the case that the permission is granted by the user. These permissions are grouped into Normal, Dangerous, Signature and SignatureOrSystem in Android. These permissions are explicitly mentioned in the AndroidManifest.xml file of APK by application developers.

Task 2 (Incident Detection) dataset contains operational log file that was captured from Unified Threat Management (UTM) of UniteCloud [18]. UniteCloud uses resilient private cloud infrastructure to supply e-learning and e-research services for tertiary students and staffs in New Zealand. Unified Threat Management is a rule based real-time running system for UniteCloud server. Each sample of a log file contains nine features. These features are operational measurements of 9 different sensors in UTM system. Each sample is labeled based on the knowledge related to the incident status of the log samples.

Task 3 (Fraud Detection) dataset is anonymised data that was unified using the highly correlated rule based uniformly distributed synthetic data (HCRUD) approach by considering similar distribution of features [19]. The detailed statistics of Task 1, Task 2 and Task 3 data sets are reported in **Table 1.**

| Task name | Total APK's | Unique APIs | Classes | Training Samples | Testing Samples |
|---|---|---|---|---|---|
| Task 1 | 61,730 | 37,107 | 2 | 30,897 | 30,000 |
| | **Total Samples** | **Features** | **Classes** | **Training Samples** | **Testing Samples** |
| Task 2 | 100,000 | 9 | 2 | 70,000 | 30,000 |
| Task 3 | 100,000 | 12 | 3 | 70,000 | 30,000 |

**Table 1** Statistics of data set

## 4.2 Hyper parameter selection for DNNs

To identify suitable parameter for DNN, we used moderately sized architecture with one hidden layer containing 128, 256, 384, 512, 640, 768, 896 and 1024 units. 2 trails of experiment are run for each parameters related to units. Each experiment is run till 200 epochs. 1024 units have shown highest 10-fold cross-validation accuracy for all use cases of cybersecurity. Thus we decided to use 1024 units for the rest of the experiments.

In order to find an optimal learning rate, we run two trails of experiment till 500 epochs with learning rate varying in the range [0.01-0.5]. The highest 10-fold cross validation accuracy was obtained by using the learning rate of 0.1. There was a sudden decrease in accuracy at learning rate 0.2 and finally attained highest accuracy at learning rates of 0.35, 0.45 and 0.45 in comparison to learning rate 0.1. This accuracy may have been enhanced by running the experiments till 1000 epochs. As more complex architectures we have experimented with, showed less performance within 500 epochs, we decided to use 0.1 as learning rate for the rest of the experiments after considering the factors of training time and computational cost.

## 4.3 DNN network topologies

The following network topologies are used in order to find an optimum network structure for our input data.

1) DNN 1 layer

2) DNN 2 layer

3) DNN 3 layer

4) DNN 4 layer

5) DNN 5 layer

For all the above network topologies, we run 2 trails of experiments. Each trail of experiment was run till 500 epochs. It was observed that most of the deep learning architectures learn the normal category patterns of input data within 600 epochs. The number of epochs required to learn the malicious category data usually varies.

The complex architecture networks required large number of iterations in order to reach the best accuracy. Finally, we obtained the best performed network topology for each use case. For Task 2 and Task 3, 4 layer DNN network performed well. For Task 1, the performance of 5 layer DNN network is good in comparison to the 4 layer DNN. We decided to use 5 layer DNN network for the rest of the experiments. 10-fold cross validation accuracy of each DNN network topology for all use cases is shown in **Table 2**.

| DNN network topology | Task Name | Accuracy |
| --- | --- | --- |
| DNN 1 layer | Task 1 | 0.712 |
| DNN 2 layer | Task 1 | 0.811 |
| DNN 3 layer | Task 1 | 0.891 |
| DNN 4 layer | Task 1 | 0.964 |
| DNN 5 layer | Task 1 | 0.978 |
| DNN 1 layer | Task 2 | 0.734 |
| DNN 2 layer | Task 2 | 0.852 |
| DNN 3 layer | Task 2 | 0.938 |
| DNN 4 layer | Task 2 | 0.991 |
| DNN 5 layer | Task 2 | 0.992 |
| DNN 1 layer | Task 3 | 0.721 |
| DNN 2 layer | Task 3 | 0.838 |
| DNN 3 layer | Task 3 | 0.912 |
| DNN 4 layer | Task 3 | 0.981 |
| DNN 5 layer | Task 3 | 0.985 |

Table 2 Summary of test results

## 4.4 Proposed Architecture

An intuitive overview of proposed DNN architecture for all use cases is shown in **Fig 1**. This contains an input layer, 5 hidden layer and output layer. An input layer contains 4896 neurons for Task 1, 9 neurons for Task 2 and 12 neurons for Task 3. An output layer contains 2 neurons for Task 1, 3 neurons for Task 2 and 2 neurons for Task 3. The details about the structure and configuration details of proposed DNN Architecture is shown in **Table 3**. The units in input to hidden layer and hidden to output layer are fully connected. DNN network is trained using the backpropogation mechanism [20]. The proposed deep neural network is composed of fully-connected layers, batch normalization layers and dropout layers.

**Fully-connected layers:** The units in this layer have connection to every other unit in the succeeding layer. That's why this layer is called as fully-connected layer. Generally, these fully-connected layers map the data into high dimension. The more the dimensions the data has the more accurate the data will be in determining the accurate output. It uses ReLU as non-linear activation function.

**Batch Normalization and Regularization:** To obviate over fitting and speed up DNN model training, Dropout (0.01) [22] and Batch Normalization [21] was used in between fully-connected layers. A dropout removes neurons with their connections randomly. In our alternative architectures for Task 1, the deep networks could easily overfit the training data without regularization even when trained on large number samples.

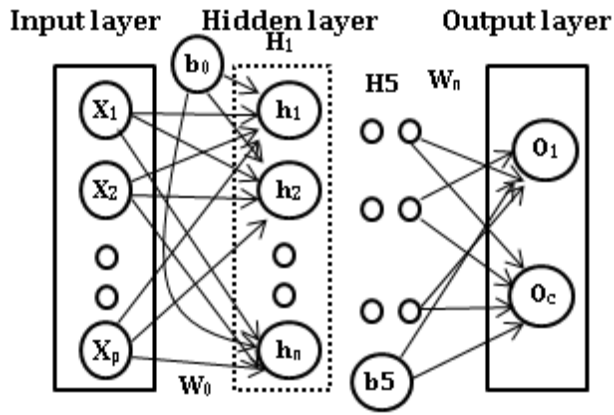

**Fig 1** Proposed deep neural architecture (DNN). All connections and units are not shown, can be considered as representative of DNN

**Classification:** For classification, the final fully connected layer follows sigmoid activation function for Task 1 and Task 2, softmax for Task 3. The fully connected layer absorb the non-linear kernel and sigmoid layer output 0 (benign) and 1 (malicious), softmax provides the probability score for each class.

The prediction loss for Task 1 and Task 2 is estimated using binary cross entropy

$$loss(pd, ed) = -\frac{1}{N} \sum_{i=1}^{N} [ed_i \log pd_i + (1 - ed_i) \log(1 - pd_i)]$$

where $pd$ is a vector of predicted probability for all samples in testing data set, $ed$ is a vector of expected class label, values are either 0 or 1.

The prediction loss for Task 3 is estimated using categorical-cross entropy

$$loss(pd, ed) = -\sum_x pd(x) \log(ed(x))$$

where $pd$ is true probability distribution, $ed$ is predicted probability distribution. We have used $sgd$ as an optimizer to minimize the loss of binary-cross entropy and categorical-cross entropy.

| Layers | Type | Output shape | Number of units | Activation function | Parameters task1&task3- (1,369,603) task2- (1,369,615) |
|---|---|---|---|---|---|
| 0-1 | Fully-connected | (None, 1024) | 1024 | ReLU | 13312 |
| 1-2 | Batch Normalization | (None, 1024) | | | 4096 |
| 2-3 | Dropout (0.01) | (None, 1024) | | | 0 |
| 3-4 | Fully-connected | (None, 768) | 768 | ReLU | 787200 |
| 4-5 | Batch Normalization | (None, 768) | | | 3072 |
| 5-6 | Dropout (0.01) | (None, 768) | | | 0 |
| 6-7 | Fully-connected | (None, 512) | 512 | ReLU | 393728 |
| 7-8 | Batch Normalization | (None, 512) | | | 2048 |
| 8-9 | Dropout (0.01) | (None, 512) | | | 0 |
| 9-10 | Fully-connected | (None, 256) | 256 | ReLU | 131328 |
| 10-11 | Batch Normalization | (None, 256) | | | 1024 |
| 11-12 | Dropout (0.01) | (None, 256) | | | 0 |
| 12-13 | Fully-connected | (None, 128) | 128 | ReLU | 32896 |
| 13-14 | Batch Normalization | (None, 128) | | | 512 |
| 14-15 | Dropout (0.01) | (None, 128) | | | 0 |
| 15-16 | Fully-connected | task1- (None, 1) task2- (None, 3) task3- (None, 1) | task1- 1 task2- 3 task3- 1 | task1&task3 -sigmoid task2- softmax | task1&task3- 129 task2-387 |
| 16-17 | Batch Normalization | (None, 1) | task1- 1 task2- 3 task3- 1 | | task1&task3- 4 task2-12 |

**Table 3** Structure and configuration details of proposed DNN Architecture

## 5 Results

We evaluate proposed DNN model against classical machine learning classifiers, on three different cybersecurity use cases. The first use case is identifying Android malware based on API information, the second use case is incident detection over unified threat management (UTM) operation on UniteCloud and the third use case is fraud detection in financial transactions. The detailed results of proposed DNN model on 3 different use cases are displayed in **Table 4**.

XGBoost is short for Extreme Gradient Boosting, where the term Gradient Boosting is proposed in the paper Greedy Function Approximation [23]. XGBoost is based on this original model. XGBoost is used for the given supervised learning problems (task1, task2 and task3), where we use the training data (with multiple features) $x_i$ to predict a target variable $y_i$. Here "multi:softmax" is used to perform the classification. After the observation and experiment, "max depth" of the tree set it as 20. 10 fold cross validation is performed to observe the training accuracy.

Except Task 1, data are loaded as it is using Pandas. The "NaN" values are replaced with 0. In Task 1 the data is represented as a term - document matrix, where the vocabulary built using the API indication numbers in train and test. The scikit-learn [24] count vectorizer used to develop the term - document matrix. On the successive representation, the data are fed to the XG Booster for prediction.

| Algorithm | Task Name | Accuracy | Precision | Recall | F-score |
|---|---|---|---|---|---|
| XGBoost | Android Malware Classification | 0.741 | 0.098 | 0.215 | 0.134 |
| XGBoost | Incident Detection | 0.997 | 0.999 | 0.998 | 0.998 |
| XGBoost | Fraud Detection | 0.916 | 0.922 | 0.916 | 0.917 |
| DNN 5 layer | Android Malware Classification | 0.940 | 0.834 | 0.868 | 0.851 |
| DNN 5 | Incident Detection | 1.00 | 1.00 | 1.00 | 1.00 |
| layer | | | | | |
| DNN 5 layer | Fraud Detection | 0.972 | 0.973 | 0.972 | 0.972 |

Table 4 Summary of test results

## 6 Conclusion

This paper has evaluated the performance of deep neural networks (DNNs) for cyber security uses cases: Android malware classification, incident detection and fraud detection. Additionally, other classical machine learning classifiers are used. In all cases, the performance of DNNs is good in comparison to the classical machine learning classifiers. Moreover, the same architecture is able to perform better than the other classical machine learning classifiers in all use cases. The reported results of DNNs can be further improved by promoting training or stacking a few more layer to the existing architectures. This will be remained as one of the direction towards the future work.

## Acknowledgement

This research was supported in part by Paramount Computer Systems. We are also grateful to NVIDIA India, for the GPU hardware support to research grant. We are grateful to Computational Engineering and Networking (CEN) department for encouraging the research.